\pdfoutput=1

\documentclass[11pt]{article}

\usepackage[]{EMNLP2023}

\usepackage{times}
\usepackage{latexsym}
\usepackage{bbm}
\usepackage{arydshln}
\usepackage{enumitem}
\usepackage{listings}
\setlist{nosep} 
\usepackage{graphicx,subcaption,ragged2e}

\usepackage{subcaption}

\usepackage{colortbl}

\definecolor{lgreen}{RGB}{73,174,137}
\definecolor{lred}{RGB}{182,49,54}
\definecolor{lorange}{RGB}{255, 128, 0}
\definecolor{lblue}{RGB}{0, 0, 255}

\colorlet{scale10}{lgreen!100}
\colorlet{scale9}{lgreen!60}
\colorlet{scale8}{lgreen!40}
\colorlet{scale7}{lgreen!20}
\colorlet{scale6}{lgreen!10}
\colorlet{scale5}{lred!5}
\colorlet{scale4}{lred!20}
\colorlet{scale3}{lred!40}
\colorlet{scale2}{lred!60}
\colorlet{scale1}{lred!75}

\colorlet{orangescale0}{lorange!0}
\colorlet{orangescale1}{lorange!10}
\colorlet{orangescale2}{lorange!20}
\colorlet{orangescale3}{lorange!30}
\colorlet{orangescale4}{lorange!40}
\colorlet{orangescale5}{lorange!50}
\colorlet{orangescale6}{lorange!60}
\colorlet{orangescale7}{lorange!70}
\colorlet{orangescale8}{lorange!80}
\colorlet{orangescale9}{lorange!90}
\colorlet{orangescale10}{lorange!100}

\colorlet{bluescale0}{lblue!0}
\colorlet{bluescale1}{lblue!10}
\colorlet{bluescale2}{lblue!20}
\colorlet{bluescale3}{lblue!30}
\colorlet{bluescale4}{lblue!40}
\colorlet{bluescale5}{lblue!50}
\colorlet{bluescale6}{lblue!60}
\colorlet{bluescale7}{lblue!70}
\colorlet{bluescale8}{lblue!80}
\colorlet{bluescale9}{lblue!90}
\colorlet{bluescale10}{lblue!100}

\newcommand{\cred}{\textsc{Cred}}
\newcommand{\bread}{\textsc{Bread}}
\newcommand{\breadrepeat}{\textsc{Bread-Repeat}}
\newcommand{\breadnoise}{\textsc{Bread-Noisy}}
\newcommand{\bleu}{\textsc{Bleu}}
\newcommand{\selfbleu}{\textsc{self-Bleu}}
\newcommand{\sacrebleu}{\textsc{SacreBleu}}
\newcommand{\chrf}{\textsc{ChrF}}
\newcommand{\chrfpp}{\textsc{ChrF++}}
\newcommand{\madlad}{\textsc{MadLad-400}}

\newcommand{\breadlabelok}{\textsc{OK}}
\newcommand{\breadlabelrepeat}{\textsc{Rep}}
\newcommand{\breadlabelboil}{\textsc{Boil}}

\usepackage{amsmath}
\usepackage[nameinlink,capitalize]{cleveref}

\usepackage[T1]{fontenc}

\usepackage[utf8]{inputenc}

\usepackage{microtype}

\usepackage{inconsolata}

\usepackage{todonotes}
\makeatletter
\newcommand*\iftodonotes{\if@todonotes@disabled\expandafter\@secondoftwo\else\expandafter\@firstoftwo\fi}  
\makeatother

\title{Separating the Wheat from the Chaff with \bread{}: \\ An open-source benchmark and metrics to detect redundancy in text}

\author{Isaac Caswell \\
  Google Research \\
  \texttt{icaswell@google.com} \\\And
  Lisa Wang \\
  Google DeepMind \\
  \texttt{wanglisa@google.com} \\\And 
  Isabel Papadimitriou \\
  Computer Science Department \\
  Stanford University \\
  \texttt{isabelvp@stanford.edu}}

\begin{document}
\maketitle
\begin{abstract}
Data quality is a problem that perpetually resurfaces throughout the field of NLP, regardless of task, domain, or architecture, and remains especially severe for lower-resource languages. A typical and insidious issue, affecting both training data and model output, is data that is repetitive and dominated by linguistically uninteresting boilerplate, such as price catalogs or computer-generated log files. Though this problem permeates many web-scraped corpora, there has yet to be a benchmark to test against, or a systematic study to find simple metrics that generalize across languages and agree with human judgements of data quality.
In the present work, we create and release \bread{}, a human-labeled benchmark on repetitive boilerplate vs. plausible linguistic content, spanning 360 languages. We release several baseline \cred{} (Character REDundancy) scores along with it,
and evaluate their effectiveness on \bread{}. 
We hope that the community will use this resource to develop better filtering methods, and that our reference implementations of \cred{} scores 
can become standard corpus evaluation tools, driving the development of cleaner language modeling corpora, especially in low-resource languages. \footnote{Our data for the BREAD benchmark and code for the CRED scores suite is at \url{https://github.com/toizzy/bread}}
\end{abstract}

\section{Introduction}

In this paper, we introduce a benchmark and propose a suite of metrics to help identify a common facet of low-quality data: repetitive boilerplate that is not reflective of natural linguistic content. Large language corpora scraped from the internet are becoming invaluable tools as self-supervised language modeling has gained prominence as a driving force of advancements in NLP \citep[][inter alia]{devlin2018bert,chowdhery2022palm,brown2020language}. 
In the case of many low-resource languages, noisy in-language data often makes up a significant proportion of any scraped corpus~\citep{kreutzer-etal-2022-quality}. Very often, this noise is in the form of repetitive boilerplate: uninteresting data without linguistic diversity, such as a long list of similar products from an e-commerce website. 
Automatically reducing repetitive boilerplate in low-resource language corpora remains an important problem to extend NLP to the thousands of languages currently underserved by language technology. 

\begin{figure}[t]
    \centering
    \includegraphics[width=0.5\textwidth]{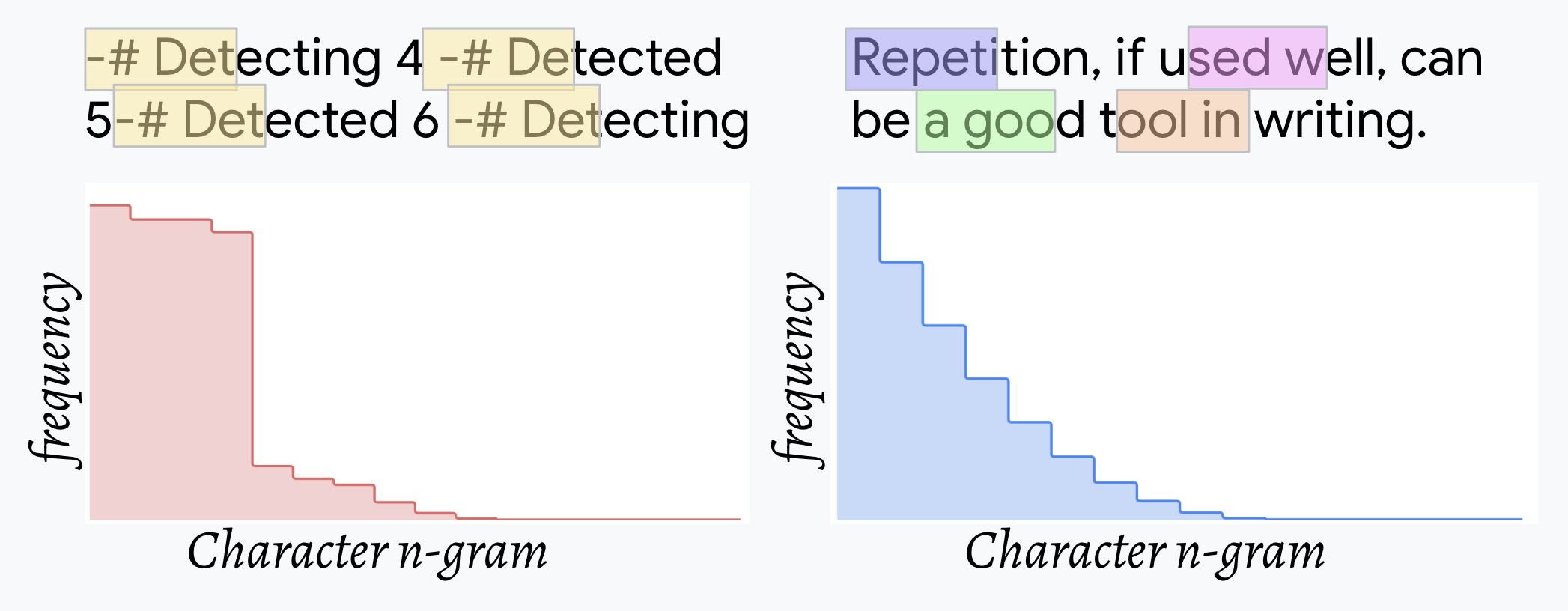}
    \caption{Character ngram based metrics compare the ngram frequency histogram between natural text and repetitive text, and assign a score of how repetitive it is. In this toy example, the character 6-gram histogram on the left is clearly distinguishable from the more natural distribution on the right. The \cred{} metrics rely on this intuition, applying simple metrics based on ngram frequency in order to detect repetitive boilerplate data in a language-agnostic manner.}
    \label{fig:diagram}
\end{figure}

To address the problem of redundant boilerplate, our contributions in the current work are two-fold:

\begin{enumerate}
    \item We release \bread{} (Boilerplate and Redundancy Evaluation on Assorted Documents), the first benchmark to measure redundancy and boilerplate in text;
    \item We test and open-source \cred{} (Character REDundancy) scores, a suite of interpretable, fast, language-agnostic metrics for detecting repetition in documents.
\end{enumerate}

Since data noise disproportionately affects low-resource languages, we only consider metrics that are language-agnostic (meaning their performance doesn't depend on any particular language). As such, we do not consider neural methods for the baselines released with \bread{}: though they are more expressive than surface-level metrics, they rely on high-quality training data, and are therefore less reliable for low-resource languages where the training data is scarce, noisy, or highly overlapping with eval or model-training data. Similarly, neural metrics struggle with interpretability and reproducibility.

The difference between a paragraph of natural text and a long, repetitive list does not depend on the source language or the particular thing that is repeating. Therefore, it is possible to build language-agnostic metrics that ignore textual features entirely, and operate purely on the token-frequency distribution. Using this intuition, we explore three ngram-based metrics: \textbf{type-token ratio (TTR)}, measuring the percentage of unique ngrams; \textbf{ngram-moment}, measuring the peakiness of the frequency distribution; and \textbf{ngram-Zipfianness}, measuring the distance from the expected frequency distribution of natural language.  

Our objective is to detect redundant language \textit{within one document}. This is different from a commonly studied problem in data quality management, where redundancy refers to a dataset containing many redundant copies of similar natural documents. We open-source the \bread{} benchmark and the \cred{} metrics, making it a replicable resource for the community.

\section{Related Work}

In the field of Data-Quality management, quality scores are used for \textit{measurement} and \textit{improvement}, and often incorporated into an iterative process \citep{wang1998product}. For NLP, there are many existing works highlighting the importance of cleaning data for training neural models \citep{khayrallah-koehn-2018-impact,junczys-dowmunt-2018-dual,denoise_nmt}. Many denoising approaches rely on classifiers \citep{chen2016a,chen2016b,D17-1155} or cross-entropy distance between models \citep{Moore2010,Axelrod2011,dynamiccds,cynlic_data_selection}, an approach often applied to data weighting and curriculum training \citep{zhang-etal-2017-boosting,wang-etal-2018-dynamic,co-curriculum}. There are neural diversity metrics, like \citet{miranda2022curse}, which uses the cosine distance between Task2Vec embeddings \citep{task2vec}, and has been applied to measure LLM output diversity \citep{lee2023beyond}.

Although data noise has always been a recognized problem, it has become a more pressing issue in recent years, as models have become more and more expressive, therefore also more capable of memorizing noise. Statistical machine translation models were more robust to data noise and tended only to benefit from bigger data \citep{goutte-etal-2012-impact} (with a few exceptions, like \citet{taghipour-etal-2011-parallel}), and works on data filtering were usually focused on improving training efficiency (for instance, ~\citet{johnson-etal-2007-improving}). Despite their generally higher performance, neural models tended to be much more sensitive to data noise~\citep{khayrallah-koehn-2018-impact}, possibly as a result of being able to memorize statistical outliers \citep{pmlr-v70-arpit17a,NEURIPS2020_1e14bfe2}. Even early versions of Paracrawl damaged MT performance \cite{junczys-dowmunt-2018-microsofts,schamper-etal-2018-rwth}, and the winners of the yearly WMT campaign tend to rely heavily on data filtering \citep{junczys-dowmunt-2018-dual,chaudhary-etal-2019-low,lu-etal-2020-alibaba,lo-joanis-2020-improving}. As a result, there have been several data filtering shared tasks in WMT \citep{koehn-etal-2018-findings,koehn-etal-2019-findings,koehn-etal-2020-findings}, and open-sourcing of various iterations of data cleaner \textsc{Bicleaner} \citep{espla-gomis-etal-2020-bicleaner,ramirez-sanchez-etal-2020-bifixer,zaragoza-bernabeu-etal-2022-bicleaner}, which use a variety of approaches, including bilingual dictionaries, random forests, and neural models.

While neural metrics or complex ensembles like \textsc{Bicleaner} are often effective, they 1) are harder to interpret; 2) may filter on artifacts like domain, rather than quality; 3) will tend only to work for languages they have explicitly been trained on; and 4) cannot be replicated between works unless a public implementation is released. For this reason, the baseline metrics released with \bread{} are simple, interpretable, surface-level metrics, that work independent of language and domain.

A token-based metric to measure the diversity and redundancy of token ngrams \textit{between} documents in a corpus (rather than within segments of one document) is \selfbleu{} \citep{zhu2018texygen}, which is based on the widely used \bleu{} score \cite{papineni2002bleu}. 
On a more granular level of character ngrams, the \chrf{} \citep{popovic2015chrf} and \chrfpp{} \citep{popovic2017chrf} metrics measure similarity between documents, correlating better with human judgement than token-level metrics like \bleu{}, especially for low-resource and highly-inflecting languages \citep{
kocmi2021ship, freitag2022stop, bapna-etal-2022-building, caswell2020language}. We follow this intuition and use character-ngram metrics. The frequency moment score defined in the present work is similar to segment-level \chrf{} applied with itself as its own reference.

Perhaps the most similar approaches to those in the present work come from a separate field, namely detecting redundancy and diversity in relational or tabular data \citep{ehrlinger2022survey}.  \citet{batista2007information} and \citet{ehrlinger2019novel} define interpretable minimality scores to measure redundancy at a schema-level for tabular data, based on cluster density, which is equivalent to the TTR in the present work.

\section{\bread{}: Dataset and Annotation}
\label{sec:dataset}

We release \bread{} (Boilerplate and Redundancy Evaluation on Assorted Documents), an expert-annotated dataset spanning 360 languages,    to tune and benchmark methods for filtering repetitive boilerplate. \bread{} consists of randomly-chosen documents from the multilingual, common-crawl-based \madlad{} dataset \citep{kudugunta2023madlad}, which are then annotated by expert NLP-practitioner annotators.

Our annotation schema consists of two high-confidence classes and two low-confidence classes. The high-confidence classes are 1) \breadlabelrepeat{}, repetitive boilerplate (N=449), and 2)  \breadlabelok{}, natural text (N=863). To keep the examples in  \breadlabelrepeat{} and  \breadlabelok{} high-confidence, we also use two low-confidence codes:  \breadlabelboil{}, for documents that are clearly non-linguistic boilerplate or noise, but are not necessarily repeating (N=499); and \textsc{unk} for where the annotator was not sure (N=3339). Documents labeled as \textsc{unk} were discarded. See Appendix Table \ref{tab:labels} for examples of each class. The examples labeled \breadlabelok{} cover 360 languages, with no individual language having more than 6 samples; the language distribution of the other three codes are harder to measure, since they are often nonlinguistic content or noisy ambiguous text.
Examples are capped at 5000 character for ease of processing.

\bread{} is split into a tune and a test set, each with 1000 documents. We propose two benchmarks, scored with F1 on the following binary prediction problems:
\begin{enumerate}
  \setlength\itemsep{0em}
    \item \textbf{\breadrepeat{}}: positive class is \breadlabelok{}; negative is  \breadlabelrepeat{}.
    \item \textbf{\breadnoise{}}: positive class is \breadlabelok{}; negative is union of \breadlabelrepeat{} and  \breadlabelboil{}.
\end{enumerate}

\section{Methods}
\label{sec:methods}

\begin{figure*}[t]
    \centering
    \includegraphics[width=0.95\linewidth]{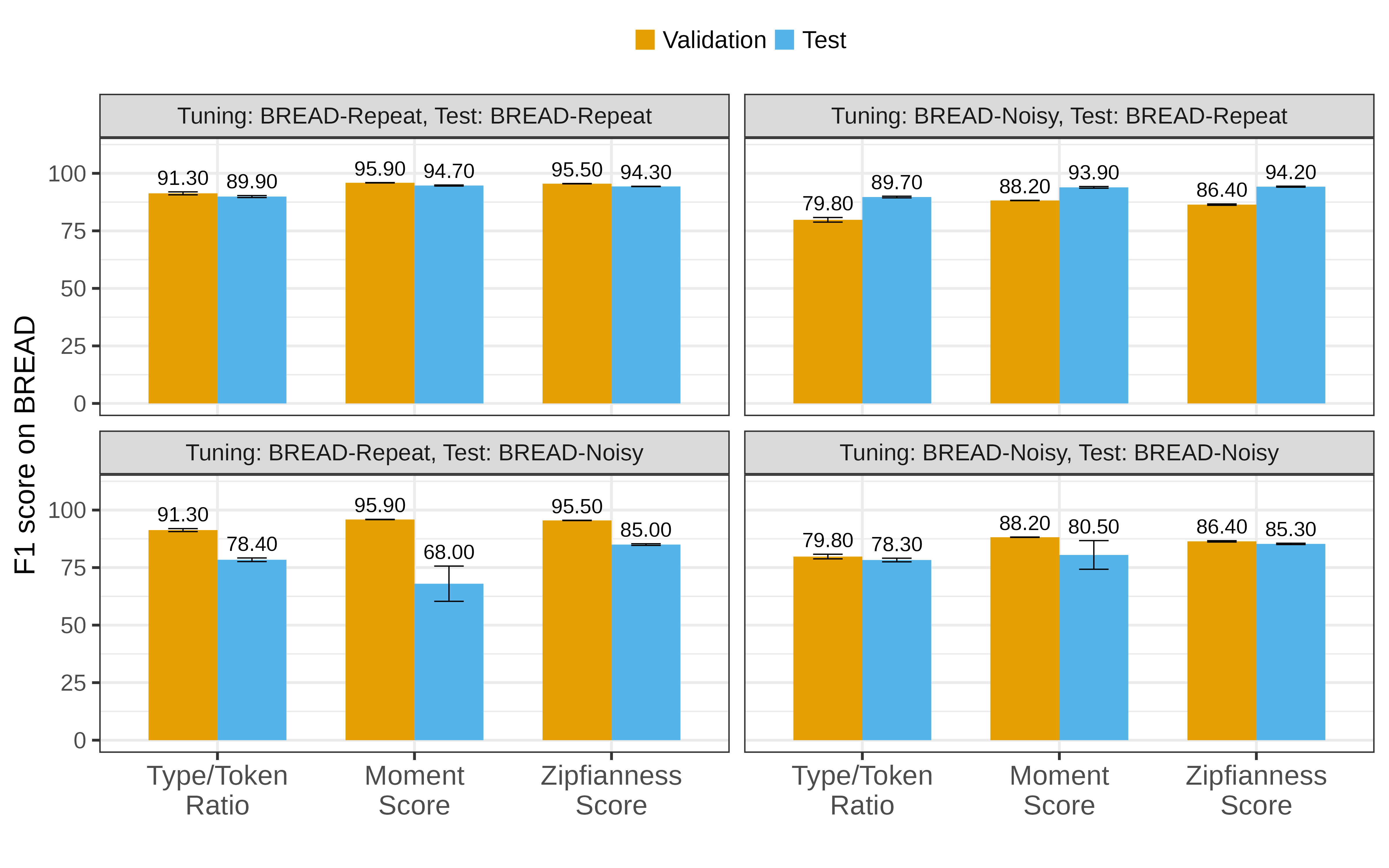} \vspace{-0.8cm}
    \caption{F1 scores for the three metrics proposed in this work, for all four combinations of tuning and testing on 
    \breadrepeat{} and \breadnoise{}. The reported values are the average of the top ten parameter settings on the tuning set. Error bars represent 95\% confidence intervals.}
    \label{fig:results_f1}
\end{figure*}

We explore three well-studied, straightforward metrics based on ngram frequency distributions and evaluate their effectiveness in the domain of measuring repetitive boilerplate. We explore both character ngrams and token ngrams, as well as combinations of the two. As with \bleu{} \citep{papineni2002bleu}, we consider using multiple $n$-gram lengths at once, and combining these scores by averaging them. By construction, all metrics assign a higher score to noisier text.

The input to all our metrics is the smoothed frequency distribution of ngrams within a document.
Distributions of ngrams tend to be noisier for shorter texts, so we apply Laplace smoothing with parameter $\lambda$, and clip the distribution with $\epsilon$-thresholding (keeping only ngrams with probability over some $\epsilon$ \citep{freitag2023epsilon}).  Let $f_n^{(i)}$ be the raw frequency of the $i$th most common $n$-gram. We define our smoothed frequency distribution as:
\begin{align}
    \tilde f_n^{(i)} \propto \left(f_n^{(i)} + \lambda\right)  \mathbbm{1} \{ f_n^{(i)} > \epsilon \}
\end{align}
This said, the authors would like to foreshadow that this distribution clipping and smoothing end up not being very important parameters for well-performing metrics, so the reader may safely ignore this and imagine that the metrics are a function of raw frequency. 

The metrics we explored are as follows:

\paragraph{TTR:} As an intuitive and well-known baseline metric, we use the Token-Type Ratio (TTR) \citep{templin1957certain}, which is the fraction of unique tokens in a document (types) over the total number of tokens. We use n-grams as tokens.

\paragraph{Frequency Moment score:} The second score we consider is the generalized moment of the frequency probability distribution, the sum of all frequencies when applying a with a nonlinearity $g(x)$. For a character n-gram with length $n$, the momet score is defined as:
\begin{align}
    m_n = \sum_i g\left(\tilde f_n^{(i)}\right) 
\end{align}
The nonlinearity $g(x)$ is a parameter we can vary to best fit out benchmark. Intuitively, setting any superlinear $g(x)$, this metric measures redundancy, or peakiness, of the ngram counts, as the score is larger when there is more weight in the head of the distribution. When $g(x) = x^k$, the score corresponds to the $k$th moment of the distribution; when $g(x) = -x\log(x)$, it corresponds to the entropy.

\paragraph{Zipfianness:} Human languages have a largely consistent word distribution: across languages, the empirical frequency of how often different words appear follows a Zipfian, or power-law distribution, where the word in frequency rank $r$ has frequency roughly proportional to $\frac{1}{r}$ \citep{zipf1936psycho, piantadosi2014zipf}. For example, in English the most common word ``the'' occurs around double the rate of the second most common word ``of''. To test whether a document is distributed like natural text, we can check whether its ngram distribution matches the empirical ngram distribution of a human language, which we estimate as a function of the n-gram length with a slight modification of the classic $\frac{1}{r}$ value (details in Appendix \ref{appendix:rgd}). \footnote{We also experiment with the empirical token distribution from a random sample of 10,000 English documents from \madlad{}-clean, and find the results to be the same (but much more painful to calculate), so for simplicity we focus only on the analytic approximation in this paper.} Therefore, we define the Zipfianness score as follows:
\begin{align}
    z_n = \sum_i d\left(\hat f_n^{(i)}, \tilde f_n^{(i)}\right)
\end{align}
Where $\hat f_i$ is the estimated frequency of the $i$th most common token, and $d(x, y)$ is a distance metric. For $d(x, y)$, we consider  $|x-y|^2, \log(|x-y|), \log^2(|x-y|)$ and $\textrm{JSD}(x, y)$. We initially also considered KL divergence (in both directions) and absolute distance, but they proved less effective.

\subsection{Compensating for Length Dependency}
\label{sec:lengthnorm}

All three of these scores are dependent on the length of the document and are all minimized when the document consists only of unique n-grams (i.e. input distribution is uniform).
Therefore, we normalize the score on a candidate document by what the score would be for a document of the same length with only unique n-grams. This leads to the interpretation of something like ``How much more redundant is this document than a natural document of the same length?". However, since natural languages are drawn from a finite and non-uniform set of symbols, the uniform distribution becomes an increasingly bad approximation of a ``natural" document as the document length increases, and leads to the reverse skew of what the length normalization was originally trying to address. To compensate for this, we introduce a simple asymptote for the number of tokens in a document, and normalize by the uniform distribution for a document with that length. This approach is chosen over the more typical approach of a fixed-width sliding window over characters, as is often done with TTR \citep{kettunen2014can}, because \bread{} has a significant range in document lengths, so we expect this approach to capture the variation in scores more cleanly. Details are in Appendix Section \ref{app:lengthnorm}.

\subsection{Grid Search}
\label{sec:grid}
Each metric is dependent on the parameters used to smooth and nonlinearize the frequency distribution, the length normalization asymptote, and the appropriate threshold when used as a classifier. Therefore, we split the dataset into a 50/50 validation/test split, and perform a grid search on the validation split, optimizing for F1 score. 
Variants of the scores optimized for different metrics are also open-sourced (\S \ref{sec:opensource});
details in Appendix \ref{appendix:grid}.

\section{Baseline Metric Results on \bread{}}

As shown in \cref{fig:results_f1}, all metrics have fairly good correlation with human judgement, even  when they are trained on the out-of-domain split of \bread{} (the off-diagonal entries). For detecting repetition alone (\breadrepeat{}; top row), both the moment score and the Zipfianness score performed about 5\% better on both tuning and test sets than TTR. When detecting both noise and boilerplate (\breadnoise{}), the difference in scores is more pronounced, with Zipfianness outperforming TTR by 9\% on the test split. The moment score, which like TTR is only able to detect redundancy but not other types of noise, barely outperforms TTR.

It is worth noting that for questions of data noise, there is a large difference between apparently close scores, if they are both close to 100. \citet{caswell2020language} note (\S \textit{Massive Class Imbalances: 99\% Accuracy Is Not Enough}) that if  a Language Identification model has a precision of 99.0, using it to generate a dataset for a typical low-resource language will yeild a dataset with just under \textit{a tenth of a percent} of sentences in the target language. Increasing this precision to 99.9\%, though under 1\% better in additive terms, is a 10x improvement in dataset precision. Keeping this in mind, we see that although we have a ways to go with better data quality scores, the improvement in noise detection from 78 F1 to 85 F1 is quite substantial!

\begin{figure*}[t]
    \centering
    \includegraphics[width=1\textwidth]{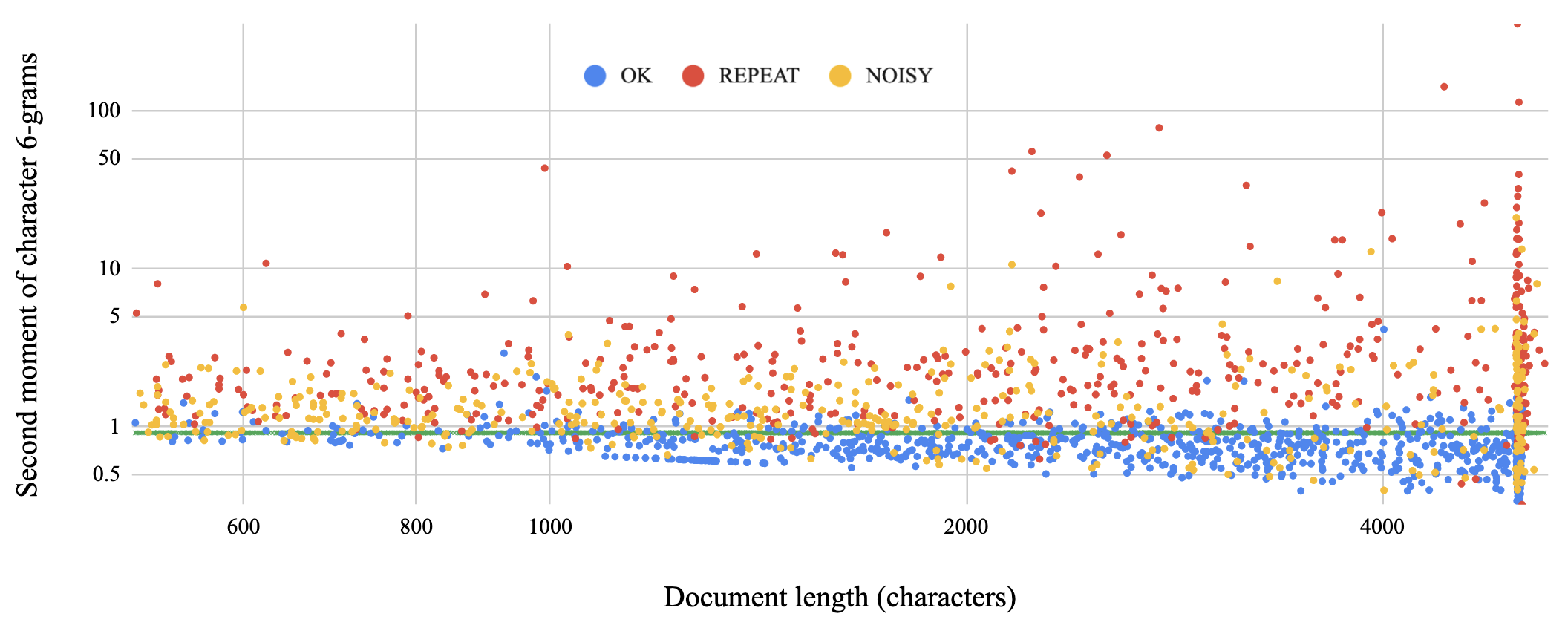}
    \caption{Moment scores on \bread{} as a function of document length, with the learned decision boundary in green, demonstrating how moment scores effectively separate noisy data from clean data along the y-axis. Each point represents a document in \bread{}, with the \breadlabelok{}  labels in blue, \breadlabelrepeat{} labels in red, and \breadlabelboil{} labels in orange. The cluster at the right reflects the truncation of \bread{} documents at 5000 characters. \label{fig:scoreviz}}
    \label{fig:diagram}
\end{figure*}

For a qualitative understanding of what scores on \bread{} look like, one can refer to Figure \ref{fig:scoreviz}, which shows the moment score as a function of length, along with the decision boundary. Details of the best hyperparameters per ngram length are given in Appendix Table \ref{tab:hparams}.

\subsection{Which Parameters Worked the Best?}

Unsurprisingly, the most important parameter was the choice of n-gram length(s). Our initial grid search went over a deep grid of different values. However, since many of these factors ended up not being very important, they led to overfitting and poor test scores. Therefore, for the final values, we re-ran the grid search with a very limited set of parameters (\S \ref{appendix:grid}). Findings from both rounds are summarized here:

\begin{itemize}
    \item {\bf n-grams:} For the purely repetition-based metrics (TTR, Moment), the  most effective n-gram length seemed to be anything of length 6-grams and up. For Zipfianness, the peak was considerably earlier, at 4-grams and 5-grams. The best single n-gram value for across all approaches would therefore be a 5-gram or 6-gram,  similar to the finding by \citet{popovic2015chrf} that 6-grams corresponded the best with human-judged quality for \chrf{}. Ensembles of different types of n-grams usually achieved slightly higher quality, but the improvements were minor.
    \item{\bf Smoothing:} There was no obvious pattern to the best smoothing value $\lambda$. 
    \item{\bf Distribution truncation:} The optimal $\epsilon$ value for  $\epsilon$-clipping was almost always 0, and the optimal $k$ for top-k clipping was almost always $\infty$. We conclude that using the full distribution is generally optimal, and omitted distribution truncation in the final grid search.
    \item{\bf Nonlinearity:} The best nonlinearity for the moment score tended to be $x^2$, corresponding nicely with the variance, though the squared entropy $(x\log(x))^2$, $x^{1.5}$, and $x^{3}$ also frequently came out on top for different settings of the other parameters. The best distance function for Zipfianness was generally the squared distance, though $\log(|x-y|)$ also performed well. 
    \item{\bf lengthnorm asymptote:} The best asymptote for the document length (used when normalizing by length; \S \ref{sec:lengthnorm}) was usually 2000.
\end{itemize}

\subsection{CRED as a Metric for Data Quality}

To validate these metrics on existing datasets and to demonstrate how they can be used to assess data quality, we report their average scores on the \madlad{} dataset. This resource is an excellent testing ground because it has both \textit{clean} and \textit{noisy} splits, and furthermore covers many very low-resource languages, where we expect more noise in the data. Results are reported in \Cref{tab:madladcleannoisy}. We make the following observations:
\begin{enumerate}
  \setlength\itemsep{0em}
    \item All three metrics agree that the \textit{noisy} split indeed has more repetitive content. This offers more evidence that our metrics are effective at detecting noise and assessing data quality.
    \item For low-resource languages (LRL), all metrics indicate that both clean and noisy splits are noisier compared to the respective splits in high-resource languages (HRL), which would align with intuition.
    \item The relative scores also allow us to make the interesting inference that the \textit{clean} split of the low-resource languages has a similar noisiness level to the \textit{noisy} split of the high-resource languages.
\end{enumerate}

\begin{table}[t]
    \centering
    \small
    \begin{tabular}{llll}
    \hline
&	TTR  &	moment &	zipf. \\
    \hline
\textsc{Mad. clean} HRL &	0.116 &	0.677 &	0.679 \\
\textsc{Mad. clean} LRL &	0.175 &	0.972 &	0.688 \\
\hline			
\textsc{Mad. noisy} HRL &	0.136 &	0.802 &	1.064 \\
\textsc{Mad. noisy} LRL &	0.189 &	1.473 &	2.063 \\
    \hline
    \end{tabular}
    \caption{Scores on the \textbf{noisy} and \textbf{clean} splits of \madlad{}, for 45 high-resource languages (HRL, >1M documents in the \textbf{clean} split) and 368 low-resource languages (LRL). All scores show more severe noise for low-resource languages, and for the \textit{noisy} split.
    }
    \label{tab:madladcleannoisy}
\end{table}

\section{Open-Sourcing}
\label{sec:opensource}

We open-source reference implementations of these metrics. Following the example of \sacrebleu{} \citep{post-2018-call}, each score has a unique signature reporting all relevant hyperparameters, so it is fully reproducible. 
In order to suit different levels of noise and different preferences of precision versus recall, we release versions of each classifier that have been tuned for F1 on a balanced version of \bread{}, as well as a version that has been tuned on the P4 score \citep{sitarz2022extending} with \bread{} upweighted so it is 75\% clean data.

\section{Conclusion}

Data quality is an evergreen problem, and as NLP is widening to a growing set of low-resource languages, where noise is a more severe problem,  the need for more interpretable metrics to asses noise becomes especially prominent. 
Recent approaches to highly multilingual technologies like NMT and LangID have reported severe noise issues for low-resource languages~\citep{caswell2020language, bapna-etal-2022-building}, and many publicly available datasets with low-resource languages in fact contain no in-language content~\citep{kreutzer-etal-2022-quality}. Nonetheless, there was heretofore no public benchmark for boilerplate and noise detection.
The present work introduces \bread{}, a multilingual, expert-annotated benchmark for detecting noise. It also investigates several interpretable, language-agnostic baseline metrics based on character ngram frequency distributions, as well as their scores on the public dataset \madlad{}. Finally, it open-sources reference implementations of several language-agnostic metrics for scoring and classifying data.

\section*{Limitations}

While the \bread{} and the metrics introduced in this paper are useful approximations, there are many forms of noise they can't detect. They can't detect poor grammar, scrambled text, translationese, toxicity, or other noise that follows a Zipfian-distribution. Furthermore they can't detect inter-example redundancy, for which a better-suited metric would be something like \selfbleu{}.

Furthermore, such a metric may not generalize well to all languages. Although the language-agnostic approach to the creation of the \bread{} eval set is constructed to work for all languages, many languages, especially those with more distinct character sets like Chinese and Japanese, may exhibit unique forms of noise or token distributions.

Finally, these metrics will tend to be less useful for shorter texts, and practitioners are cautioned against using them on sentence-level data.

\section*{Ethics Statement}

We introduce a benchmark dataset and scoring mechanisms for improving the quality of low-resource language corpora. Like any metrics based on surface-level features, our metrics are coarse and do not reflect the subtleties of different languages. We propose for our CRED scores to be used in a battery of data quality evaluation methods.

\section*{Acknowledgements}

The authors would like to thank the wonderful Ithaka Co-Operative House, which enabled this collaboration.  
This research was funded in part by NSF award number IIS-2128145.

\bibliography{anthology,custom}
\bibliographystyle{acl_natbib}

\appendix

\clearpage
\section{Length normalization details}
\label{app:lengthnorm}

As mentioned in Section \ref{sec:lengthnorm}, these simple metrics have a dependency on the length of the document, which is undesirable. Therefore, we normalize them by dividing by their minimum possible value for a document of that length, which is achieved on the uniform distribution. (The maximizing value, achieved by the one-hot distribution, grows very quickly only seemed to add noise.)

\subsection{Moment}

\begin{figure*}
\captionsetup[subfigure] {justification=Centering}

\begin{subfigure}[t]{0.5\textwidth}
    \includegraphics[width=\textwidth]{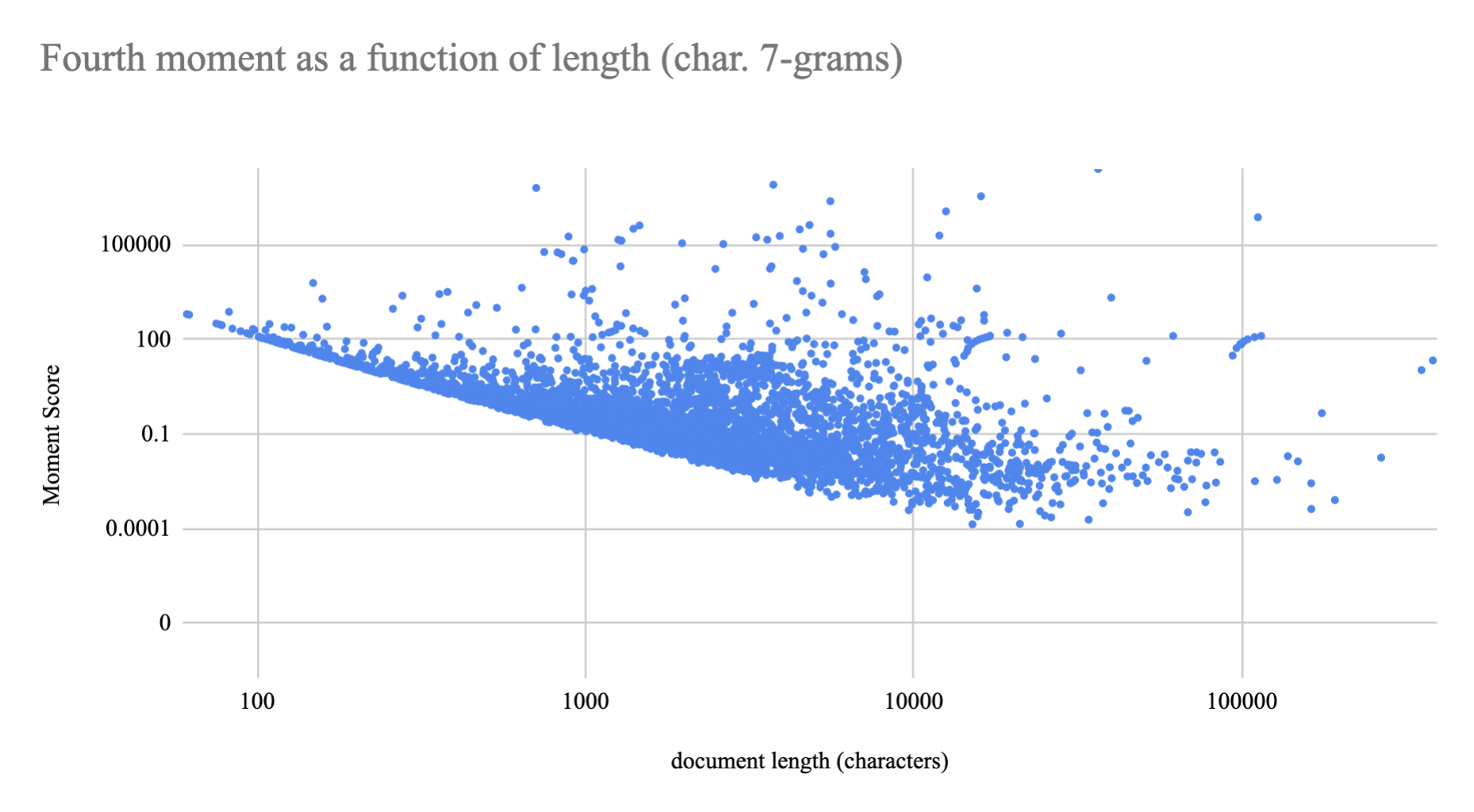}
    \caption{Distribution of the fourth moment score on character 7-grams (found to be the most effective for \breadrepeat{}) on relatively clean samples of seven languages, as a function of document length. \label{fig:lengthnorma}}
\end{subfigure}\hspace{\fill} 
\begin{subfigure}[t]{0.5\textwidth}
    \includegraphics[width=\linewidth]{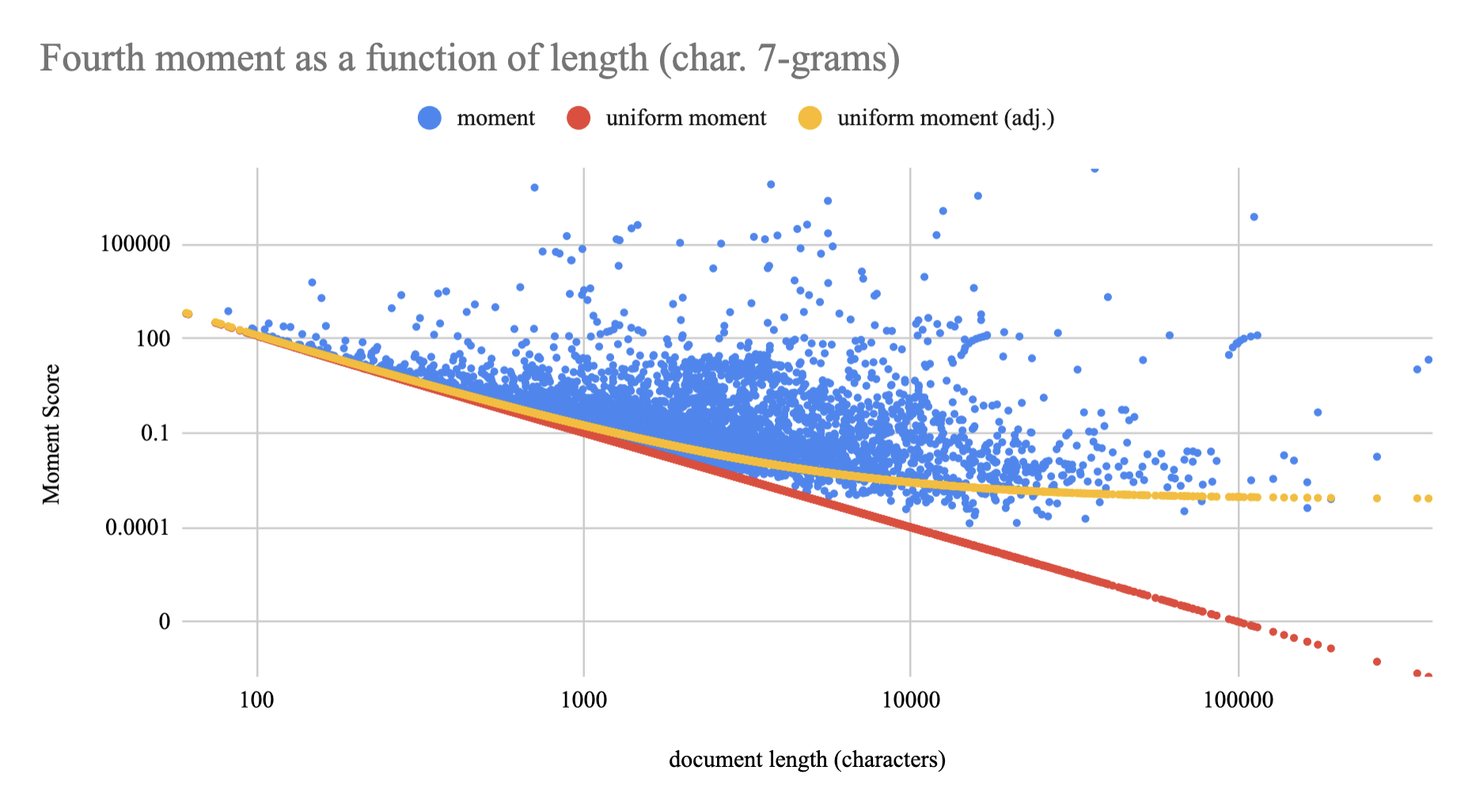}
    \caption{The moment of the uniform distribution (red) and the adjusted uniform distribution (yellow), where the latter simply interpolates between the number of n-grams in a document and a max-ngram value of 5000\label{fig:lengthnormb}}
\end{subfigure}

\bigskip 
\begin{subfigure}[t]{0.5\textwidth}
    \includegraphics[width=\linewidth]{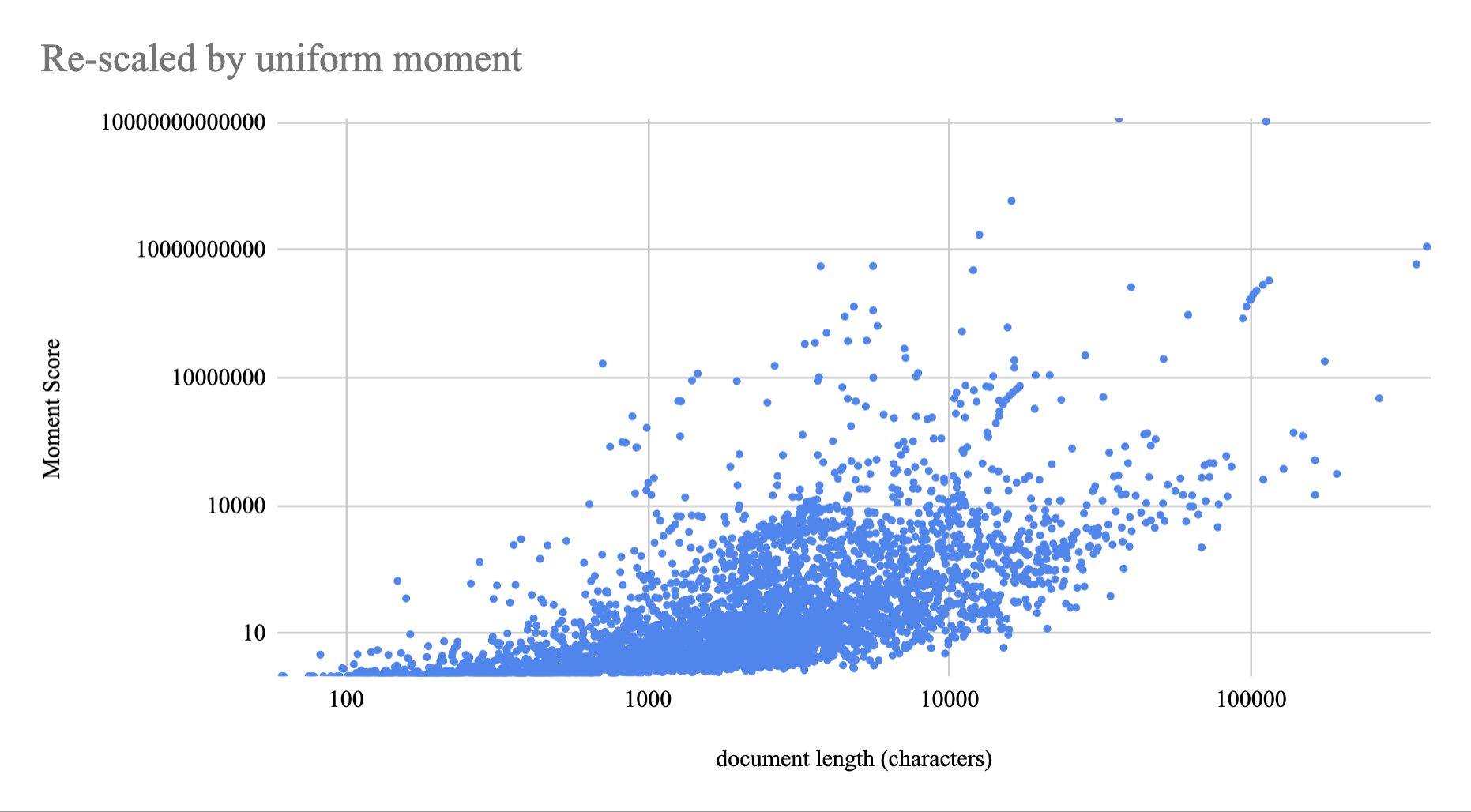}
    \caption{Moment score normalized by the uniform moment. It is apparent that the score is still length-dependent.\label{fig:lengthnormc}}
\end{subfigure}\hspace{\fill} 
\begin{subfigure}[t]{0.5\textwidth}
    \includegraphics[width=\linewidth]{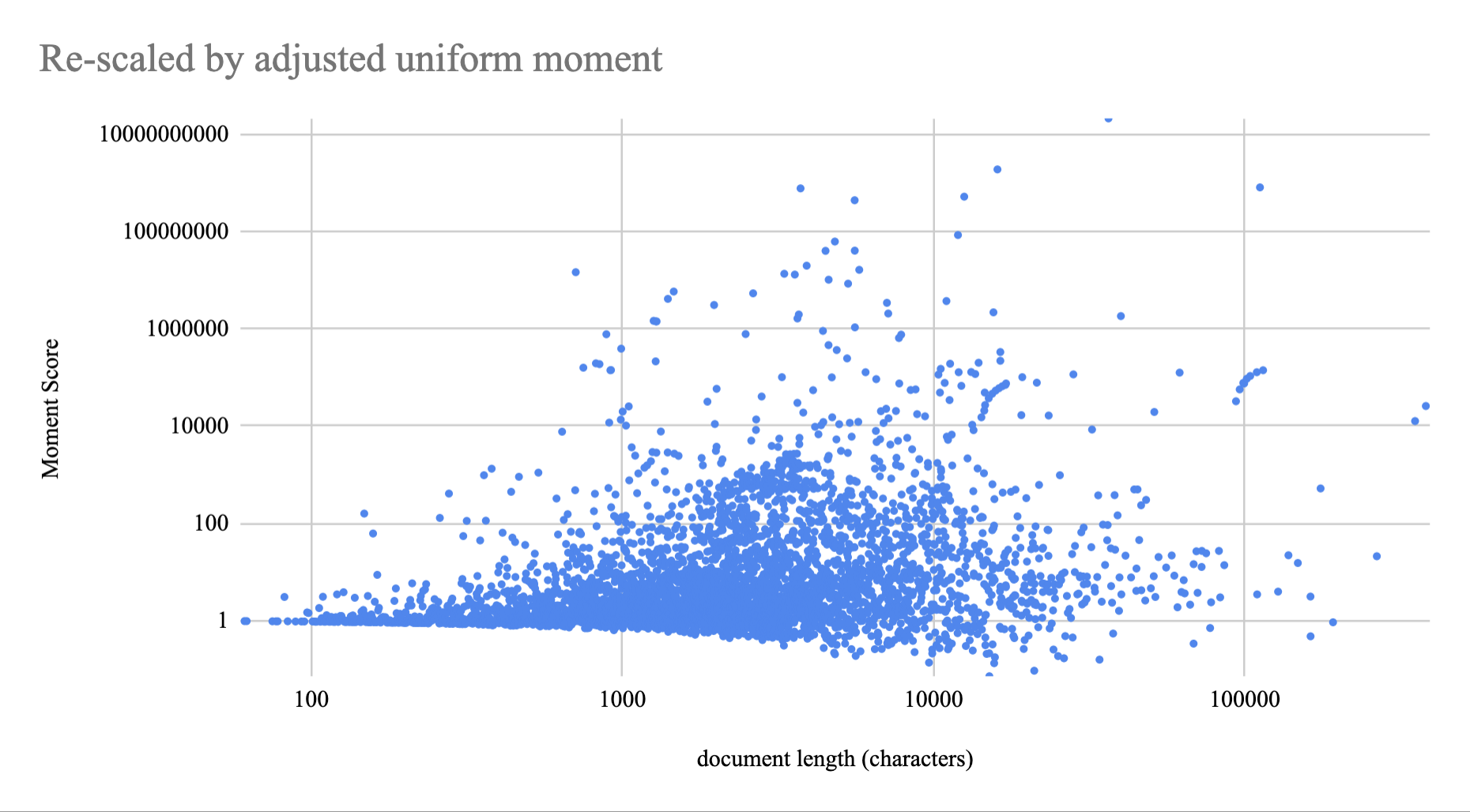}
    \caption{Moment score normalized by the adjusted uniform moment. The length dependency is much less.\label{fig:lengthnormd}}
\end{subfigure}

\caption{Length normalization for the moment score}

\end{figure*}

The distribution of moment scores on a sample of filtered, web-mined text across a variety of languages\footnote{Equal mix of Arabic, English, Finnish, German, Russian, Swahili, and Turkish} can be seen in Figure \ref{fig:lengthnorma}. There is a clear lower bound on this distribution, which corresponds to a uniform token distribution. In Figure \ref{fig:lengthnormb}, the distribution is plotted alongside the score on the uniform distribution (in red) and in Figure \ref{fig:lengthnormc}, the moment scores are shown when normalized by the uniform distribution. It is clear that this is a poor fit for longer documents, where the uniform distribution is more unlikely, and indeed (in the case of a finite alphabet) impossible. Therefore, we introduce an asymptote on the length of the document. For a document with true length $n$, we instead calculate the uniform distribution moment on a scaled length $\tilde n = \frac{n*\alpha}{n+\alpha}$, for some asymptote $\alpha$. The yellow line in Figure \ref{fig:lengthnormb} shows the uniform distribution on $\tilde n$ with an asymptote of $\alpha = 5000 $, and Figure \ref{fig:lengthnormd} demonstrates that after dividing by this, the length dependency, at least when it comes to the lower bound, has nicely flattened out.

\subsection{Zipfianness}

We normalize the Zipfianness score in the same way as the moment score, namely by the score on the uniform distribution, with some asymptote parameter $\alpha$.

\section{Grid Search}
\label{appendix:grid}

Each metric is dependent on the parameters used to smooth and nonlinearize the frequency distribution (Section \ref{sec:methods}). Furthermore, in order to use such a metric as a classifier for whether text is noisy or not, an appropriate threshold is needed as the decision boundary. Therefore, for each metric, we carry out a grid search over its possible hyperparameters. We split the \bread{} dataset into a 50/50 tune/test split, and perform the grid search on the tune split. \footnote{A train split per se is not necessary, as we are not training any models.} The hyperparameter ranges initially explored were as follows:

\vspace{10px}

\textbf{Grid Search 1:}

\begin{itemize}
    \item \textbf{ngrams:} we explore every contiguous combination of ngrams from character 2-grams to character 10-grams. We also explore token 1-grams and 2-grams, and combinations of token 2-grams with character 5- and 6-grams, as in \chrfpp{}.
    \item \textbf{$\epsilon$ values}: we cover the range of [0, 0.01]
    \item \textbf{$k$ values}: we cover the range of [2, 1024], as well as no top-k filtering
    \item \textbf{smoothing}: we cover the range of [0, 2].
    \item \textbf{nonlinearities}: These vary by method and are described along with each method.
\end{itemize}

\vspace{14px}
However, given the small size of the tuning metric, this led to severe overfitting. Based on analysis of which parameters were or were not very important, we re-did the final, simpler grid search:

\vspace{10px}

\textbf{Grid Search 2 (constrained):}

\begin{itemize}
    \item \textbf{ngrams:} we explored only sets of one to two ngram values at once, for instance a mixture of 4-grams and 5-grams, but not larger sets like in the first gridsearch. For multiple-ngram settings we looked at contiguous lengths as well as skip-2 lengths. We explored character 1-grams to character 10-grams.
    \item \textbf{$\epsilon$ values}: we did not do epsilon truncation.
    \item \textbf{$k$ values}: we did not perform top-k filtering
    \item \textbf{smoothing}: we  only explored 0 and 1.
    \item \textbf{nonlinearities}: We limited ourselves to $x^{1.5}, x^2, x^3$ for the moment, and $|x-y|^2, \log(|x-y|), \log^2(|x-y|), \textrm{JSD(x, y)}$ for Zipfianness.
\end{itemize}

We optimized the grid search with the F1 metric. The choice of the optimization metric is inherently dependent on the data balance, and a one-size-fits-all solution is not possible; as such, though this is the metric explored in this paper, variants of the scores optimized for different metrics are open-sourced (See Section \ref{sec:opensource}).

\section{Dataset Classes and Examples}
Several examples of documents annotated with different classes from the \bread{} dataset are given in Table \ref{tab:labels}.

\begin{table*}[t]
  \centering
  \small
    \begin{tabular}{lll}
    \hline
Class &	Description & Example \\
    \hline
OK & natural text & Alokba den Sangremer Sensaksem | Tir Yimyim \\
& &By nungsang on November 28, 2017 Comments Off on Alokba den Sangremer Sensaksem \\
& &Sangremer : Alokba, nenok ashiakang kijong tepenjem kibong, Okolai nabo tulura ta \\
& &meteta lir, saka nü kinungtsü indangang junga memetet. La kechi inyaker-aka? \\
& &Tenünga shiba aka? \\
& &Alokba : Oko, Labo mapangshia polashia tzüwa awaba dak alaka kecha balaka \\
& & meinyakerako, la tenünga Tzüwala, süra ner tantsüa kechi inyaker? \\
\hline
& & Kipirde basylan türkmen migrantlar \\
& &Türk polisiýasy we migrant. \\
& &Kipriň demirgazyk böleginiň metbugatynda soňky wagtlarda türkmen migrantlary \\
& &barada köp maglumat çykyp başlady. Diňe soňky birnäçe günüň dowamynda ol ýerde\\
& &birnäçe türkmen zähmet migrantlarynyň ogurlykda aýyplanyp, suda çekilip, soňra-da \\
& &wagtlaýynça tussag edilendigi habar berilýär. \\
& &Belli bolşy ýaly Türkmenistan garaşsyzlygyny alandan soň Türkiýe türkmen zähmet \\
& &migrantlarynyň esasy ýykgyn edýän ýurtlarynyň birine öwrüldi. Türkiýedäki türkmen \\
& &zähmet migrantlary barasynda türk metbugatynda yzygiderli maglumatlar çap edilýär. \\
& &Ýöne indi Türkiýeden Kipriň demirgazyk bölegine gidip işleýän türkmenistanly zähmet \\
& &migrantlary barada hem metbugatda çap edilýän maglumatlar köpelýär. \\
    \hline
    \hline
REP & repetitive & Shabir May 13, 2019 at 8:24 PM \\
& boilerplate &Shabir May 13, 2019 at 8:27 PM \\
& &Shabir May 13, 2019 at 8:28 PM \\
& &Do visit the site Eduassam jobs in Assam \\
& &tridip May 31, 2019 at 8:24 PM \\
& &golam June 12, 2019 at 10:48 PM \\
\hline
& &3.6 miles 18° 2020-01-13 12:16:54 \\
& &3.7 miles 181° 2020-01-18 14:04:11\\
& &3.7 miles 181° 2020-01-19 19:29:48\\
& &3.8 miles 235° 2020-01-20 19:43:23\\
& &Stations qui ont entendu WA1PLE-13 directement par radio – \\
& & 2020-012019-122019-112019-102019-092019-082019-07 \\
& &1 2020-01-14 03:19:07 2020-01-14 03:19:07 FN42JD > \\
& &FN31ST 67.3 miles 250° 2020-01-14 03:19:07\\
& &54 2020-01-09 00:45:56 2020-01-19 06:52:58 FN42JD >\\ 
& & FN42BF 32.4 miles 282° 2020-01-19 06:52:58\\
& &1 2020-01-15 01:20:00 2020-01-15 01:20:00 FN42JD > \\
& & FN33TA 84.8 miles 317° 2020-01-15 01:20:00\\
    \hline
    \hline
BOIL & boilerplate & jasa service rolling door murah: jasa service kunci rolling \\
& but not &murah jakarta selatan,utara,pusat,slipi,sunter, tangerang.\\
& repeating &jasa service kunci rolling murah jakarta selatan,utara,pusat,slipi,sunter, tangerang. \\
& &Diposting oleh ardicom di 18.53 \\
\hline
& &E5500/6500 68" Cabinet 4U Rack Mount Kit -- Sun Parts from AnySystem.com. \\ 
& & X9674A 595-5540 For pricing and availability, please call 201-445-3122 \\
& &or email sales@anysystem.com . \\
& & AnySystem - Home / X9674A 595-5540 E5500/6500 68" Cabinet 4U Rack Mount Kit  \\
& & -- Sun Parts E5500/6500 68" Cabinet 4U Rack Mount Kit -- Sun Parts from AnySystem.com. \\
\hline
    \end{tabular}
    \caption{\bread{} Dataset classes and corresponding examples. Note that some examples are excerpts from longer documents.}
    \label{tab:labels}
\end{table*}

\clearpage
\section{Zipf Approximation via Random Gradient Descent}
\label{appendix:rgd}

We initially calculated the empirical Zipf distribution from a linguistically diverse set of data. However, this was cumbersome to deal with, since we needed a value for every n-gram length and for every n-gram index, leading to a 20x10000 table. Although the approximation of $f_r \propto \frac{1}{r^b}$, for the 1-indexed rank of a token $r$ and some exponent $b$, is an ok approximation, it is known to be fairly poor near the edges of the distribution. Therefore, we used the following algorithm to determine a better approximation, which we call Random Gradient Descent (RGD). The basic approach is to perturb a point randomly until the loss function improves, and then follow that direction in the parameter space until the loss stops decreasing, and alternate doing these two steps until convergence. In pseudocode, this algorithm looks like this:

\begin{lstlisting}[language=Python,
    basicstyle=\small
]

def rgd(initial_args, loss_fn,
lr=0.01,
branch_n=10,
max_steps=10000,
max_attempts=10):
  total_steps = 0
  best_args = initial_args.copy()
  n_failed = 0
  cur_loss = loss_fn(best_args)
  initial_loss = cur_loss
  it = 0
  while True:
    it += 1
    if total_steps >= max_steps: break
    total_steps += branch_n
    branch, branch_grad, branch_loss =
    get_best_branch(best_args, loss_fn,
    lr, branch_n)
    if branch is None:
      # This means that no branch
      # improved on the best args.
      # As a result, there is no 
      # gradient to follow.
      n_failed += 1
      if max_attempts 
        and n_failed >= max_attempts:
        break
      continue
    cur_loss = branch_loss
    n_failed = 0
    best_args, follow_steps, follow_loss = 
    follow_grad(branch, branch_grad,
    loss_fn)
    total_steps += follow_steps
    cur_loss = follow_loss
  return best_args, cur_loss, total_steps





  
def get_best_branch(args, loss_fn,
                    lr, branch_n):
  """Look at branch_n random points 
  around args. Return the one with 
  the lowest loss, and if none of them
  decreases the loss, return None's.
  """
  cur_loss = loss_fn(args)
  pool_args = [(args, lr, loss_fn)
  for _ in range(branch_n)]
  
  with Pool() as p:
    result = 
    p.map(eval_branch, pool_args)
  branches, losses, grads = zip(*result) 
  best_loss = min(losses)
  if best_loss >= cur_loss:
    return None, None, None
  i = losses.index(best_loss)
  return branches[i], grads[i], best_loss
  
def follow_grad(args, grad, loss_fn, 
max_flat=20):
  """Follow the gradient grad
  until the loss stops improving.
  Guaranteed never to make the
  loss worse; might not change it.
  """
  cur_loss = loss_fn(args)
  initial_loss = cur_loss
  best_args = args.copy()
  n_flat = 0
  total_steps = 0
  while True:
    new_args = take_step(best_args, grad)
    new_loss = loss_fn(new_args)
    if new_loss > cur_loss: break
    elif new_loss == cur_loss:
      n_flat += 1
      if n_flat >= max_flat:
        break
    elif new_loss < cur_loss:
      total_steps += 1 + n_flat
      n_flat = 0
      best_args = new_args.copy()
      cur_loss = new_loss
  return best_args, total_steps, cur_loss
\end{lstlisting}

The literature is certainly rich with better and  subtler ways to find a good approximation, but this method yielded an approximation that performed as well as the empirical Zipf distribution with our methods. The approximation we found with this method, and which we used in the main paper, is as follows, for the $r$th most common character n-grams of length $n$ :

\begin{align*}
b &= 6.809 *  (r + 2.768)^{-1.487} + 0.527 \\
s &= 0.107 * (n + 12.0147)^{-12.654} + 0.0139 \\
  f_{r}^n &= s\frac{1}{r^b}
\end{align*}

\begin{table*}[t]

\parbox{.45\linewidth}{
    \centering
    \begin{tabular}{ll|lllll}
    \hline
score & 	n  &	Tune  & Test   &	$\alpha$ &	nl &	$\lambda$ \\
\hline					
TTR &  1 &  82.6 & 82.0 & NA & NA & NA \\
TTR &  2 &  82.8 & 82.4 & NA & NA & NA \\
TTR &  3 &  83.9 & 83.4 & NA & NA & NA \\
TTR &  4 &  86.3 & 85.1 & NA & NA & NA \\
TTR &  5 &  87.9 & 87.5 & NA & NA & NA \\
TTR &  6 &  89.4 & 89.2 & NA & NA & NA \\
TTR &  7 &  90.8 & 90.2 & NA & NA & NA \\
TTR &  8 &  91.8 & 90.4 & NA & NA & NA \\
TTR &  9 &  92.4 & 90.7 & NA & NA & NA \\
TTR &  10 &  92.5 & 90.1 & NA & NA & NA \\
\hline      
mmt. &  1 &  83.3 & 82.9 & $\infty$ & $x^{1.5}$ & 1 \\
mmt. &  2 &  84.4 & 84.6 & 2k & $x^{1.5}$ & 0 \\
mmt. &  3 &  87.4 & 88.0 & 2k & $x^{1.5}$ & 0 \\
mmt. &  4 &  92.3 & 91.7 & 2k & $x^{1.5}$ & 1 \\
mmt. &  5 &  95.2 & 93.7 & 2k & $x^{1.5}$ & 0 \\
mmt. &  6 &  95.8 & 94.9 & 2k & $x^{1.5}$ & 0 \\
mmt. &  7 &  95.8 & 94.3 & 2k & $x^{3}$ & 1 \\
mmt. &  8 &  95.8 & 94.6 & 2k & $x^{3}$ & 1 \\
mmt. &  9 &  95.4 & 94.7 & 2k & $x^{3}$ & 1 \\
mmt. &  10 &  95.1 & 94.5 & 5k & $x^{2}$ & 0 \\
\hline      
Zipf &  1 &  83.1 & 82.2 & 2k & $\log(x)$ & 1 \\
Zipf &  2 &  84.7 & 84.0 & $\infty$ & $\log(x)$ & 0 \\
Zipf &  3 &  90.1 & 89.7 & 2k & JSD & 1 \\
Zipf &  4 &  94.7 & 93.7 & 2k & $x^{2}$ & 0 \\
Zipf &  5 &  95.5 & 94.3 & 5k & $x^{2}$ & 0 \\
Zipf &  6 &  94.5 & 93.4 & $\infty$ & $x^{2}$ & 0 \\
Zipf &  7 &  93.5 & 92.8 & 2k & $x^{2}$ & 0 \\
Zipf &  8 &  92.7 & 92.3 & 2k & $x^{2}$ & 0 \\
Zipf &  9 &  91.8 & 91.3 & 2k & $x^{2}$ & 0 \\
Zipf &  10 &  91.2 & 90.7 & 2k & $x^{2}$ & 0 \\
\hline					
\end{tabular}					
\caption{Eval on \breadrepeat{}}					
\label{tab:hparams}					
}					
\hfill					
\parbox{.45\linewidth}{					
\centering					
\begin{tabular}{ll|lllll}					
\hline					
score &	n &	Tune &Test & $\alpha$ &	nl &	$\lambda$ \\	
\hline					
TTR &  1 &  70.7 & 70.3 & NA & NA & NA \\
TTR &  2 &  70.8 & 70.5 & NA & NA & NA \\
TTR &  3 &  71.6 & 71.3 & NA & NA & NA \\
TTR &  4 &  73.6 & 72.8 & NA & NA & NA \\
TTR &  5 &  75.2 & 74.8 & NA & NA & NA \\
TTR &  6 &  77.3 & 76.7 & NA & NA & NA \\
TTR &  7 &  79.0 & 77.7 & NA & NA & NA \\
TTR &  8 &  80.6 & 79.1 & NA & NA & NA \\
TTR &  9 &  81.6 & 79.7 & NA & NA & NA \\
TTR &  10 &  81.9 & 79.6 & NA & NA & NA \\
\hline
mmt. &  1 &  71.3 & 49.5 & $\infty$ & $x^{1.5}$ & 1 \\
mmt. &  2 &  72.7 & 56.4 & 2k & $x^{1.5}$ & 0 \\
mmt. &  3 &  75.4 & 58.5 & 2k & $x^{1.5}$ & 0 \\
mmt. &  4 &  81.2 & 57.7 & 2k & $x^{1.5}$ & 1 \\
mmt. &  5 &  86.2 & 66.9 & 2k & $x^{1.5}$ & 1 \\
mmt. &  6 &  88.0 & 79.6 & 2k & $x^{2}$ & 1 \\
mmt. &  7 &  88.0 & 87.6 & 2k & $x^{2}$ & 0 \\
mmt. &  8 &  88.3 & 87.9 & 2k & $x^{2}$ & 0 \\
mmt. &  9 &  88.1 & 87.2 & 2k & $x^{2}$ & 0 \\
mmt. &  10 &  87.5 & 64.1 & 5k & $x^{3}$ & 1 \\
\hline
Zipf &  1 &  71.0 & 61.3 & 2k & $\log(x)$ & 1 \\
Zipf &  2 &  73.0 & 72.4 & $\infty$ & $\log(x)$ & 0 \\
Zipf &  3 &  78.6 & 48.7 & 2k & JSD & 1 \\
Zipf &  4 &  86.2 & 85.5 & 2k & $x^{2}$ & 0 \\
Zipf &  5 &  86.0 & 85.4 & 2k & $x^{2}$ & 0 \\
Zipf &  6 &  84.2 & 84.4 & 2k & $x^{2}$ & 0 \\
Zipf &  7 &  82.5 & 82.4 & 2k & $x^{2}$ & 0 \\
Zipf &  8 &  81.4 & 81.2 & 2k & $x^{2}$ & 0 \\
Zipf &  9 &  80.1 & 79.8 & 2k & $x^{2}$ & 0 \\
Zipf &  10 &  80.0 & 72.0 & $\infty$ & $\log(x)$ & 1 \\
\hline					
\end{tabular}					
\caption{Eval on \breadnoise{}}					
\label{tab:hparams}					
}					
\caption{F1 and Parameters of the scores that maximized the tune F1 on \breadrepeat{} and \breadnoise{}, for all combinations of character n-gram length and score type. The parameters in question are the length-normalization asymptote $\alpha$, the nonlinearity \texttt{nl}, and the Laplace smoothing parameter $\lambda$. 	Perhaps the most interesting thing to note is when the tune/test F1 scores as a function of ngram size: for the two metrics that only detect repetition (TTR and Moment), larger ngrams are generally better, whereas for Zipfianness, utility peaks around 5.					\label{tab:hparams}				
}						
\end{table*}

\end{document}